# Domain-Adversarial Adaptation from MURA to In-House Radiographs for Improved Adult Elbow Fracture Classification

M. Bolhassani[1], W. Furtado[2], D. Duvall[2], MJ Negahdar[2], J. Joshi[2], S. Contractor[2], A. Amini[1]
[1]Department of Electrical and Computer Engineering, University of Louisville, KY
[2] Radiology Department, University of Louisville, Louisville, KY

## I. Description of Purpose:

We studied domain shift in adult elbow fracture classification from radiographs when training on a large public dataset [1] and deploying on a small, institution-specific dataset (UofL). We implemented a domain-adversarial neural network (DANN) [2] to align feature distributions across source (MURA) and target (UofL) domains while optimizing fracture classification. A pretrained convolutional backbone was adapted to grayscale inputs; the task head predicts fracture presence, and a gradient-reversal-based domain head encourages domain-invariant features via adversarial learning. We compared four settings: (1) Source-only, (2) Target-only (limited labels), (3) Naive merge (pooled training), and (4) DANN (MURA to UofL). Evaluation on UofL test data uses AUC-ROC, and accuracy. DANN outperformed Source-only and Naive merge baselines by reducing domain bias without overfitting to the small target set. Ablations showed that progressive adversarial weight, dataset-size-aware domain loss, and domain-balanced sampling improved stability and target performance. Results demonstrated that domain alignment is a practical path to robust clinical generalization from public to local data.

## II. Introduction

Performance of deep learning models trained on public medical imaging datasets often degrade on local clinical data due to differences in scanners, protocols, populations, and annotation practices (domain shift). For elbow fracture classification, MURA [1] provides much of the scale required for representation learning, but our target deployment is an in-house cohort (UofL) whose distribution may differ. Standard remedies like naive pooling or source-only pretrain which then finetuning on target, do not explicitly close the domain gap. Domain-adversarial training (DANN) [2] tackles this by learning features that are both discriminative for the clinical task and indistinguishable across target and source domains. Prior studies show that adversaria domain adaptation can improve generalization in medical data across institutions and acquisition conditions. [3] aligned prostate histopathology whole-slide images with an adversarial objective and reported gains under substantial inter-laboratory variability, while [4] proposed CDE-Net, a class–decision-boundary–aware, dynamically expanding network that preserves source-domain information while progressively adapting to the target distribution; by combining clustering loss with conditional-entropy minimization achieving large average AUC improvements across chest X-rays, intracranial hemorrhage, and mammography datasets. To minimize risk of anatomically implausible alterations, we do not perform any pixel-level style transfer; all adaptation occurs in feature space via gradient reversal, leaving raw radiographs unchanged. Model selection is target-focused; we choose the best selected model with the highest ROC-AUC on a hold-out UofL validation split. We adapt DANN for our setting by (a) handling grayscale radiographs, (b) using pretrained backbones suitable for medical images, and (c) adding training stability improvements, for small target datasets. Our contributions are as follows:

1. A DANN pipeline that aligns MURA (source) and UofL (target) for elbow fracture classification with grayscale-aware backbones.
2. A stable training recipe for medical domain adaptation (DA): progressive adversarial weighting, dataset-size-aware domain loss, and early stopping on target-validation ROC-AUC.
3. A comprehensive evaluation against Source-only, Target-only, and Naive-merge baselines.
4. Qualitative analysis of alignment via PCA and Grad-CAM to ensure task-relevant, not spurious, invariances.

Among domain-adaptation paradigms, we selected DANN because it explicitly suppresses domain-specific variance while preserving task-discriminative cues, requires no target labels beyond a small validation split, and avoids image-level style transfer that can distort anatomy.

## III. Data

Following approval from the University of Louisville Institutional Review Board, we obtained elbow radiographs captured in multiple views from a cohort of 379 subjects with Each subject's radiographs included two views: AP and Lateral. Recognizing the variability in the difficulty of diagnoses ranging from simple to challenging, the selected data encompassed a spectrum of diagnosis difficulty levels. The more complex cases often contain subtle signs of fracture, which only a highly trained radiologist can identify. Figure 1(a) displays AP and lateral view examples of both a straightforward case and a challenging case [5].

UofL Dataset: Training and test sets from the dataset were randomly partitioned as demonstrated in Table 1, ensuring an equal distribution of samples across different levels of difficulty. We devised a semi-automated cropping procedure to have all training samples focused on the joint area, eliminating background regions. We also enhanced the contrast of grayscale images using the Contrast Limited Adaptive Histogram Equalization (CLAHE) technique [6] and normalized the intensities between 0 and 1. Subsequent to all preprocessing steps, all data had dimension (640×640×1) for both training images and their corresponding masks. For data distribution across training, evaluation, and testing, we first allocated 10 percent of the data from each difficulty level for the test set. The remaining samples were then divided into training and evaluation sets in an 80:20 ratio.

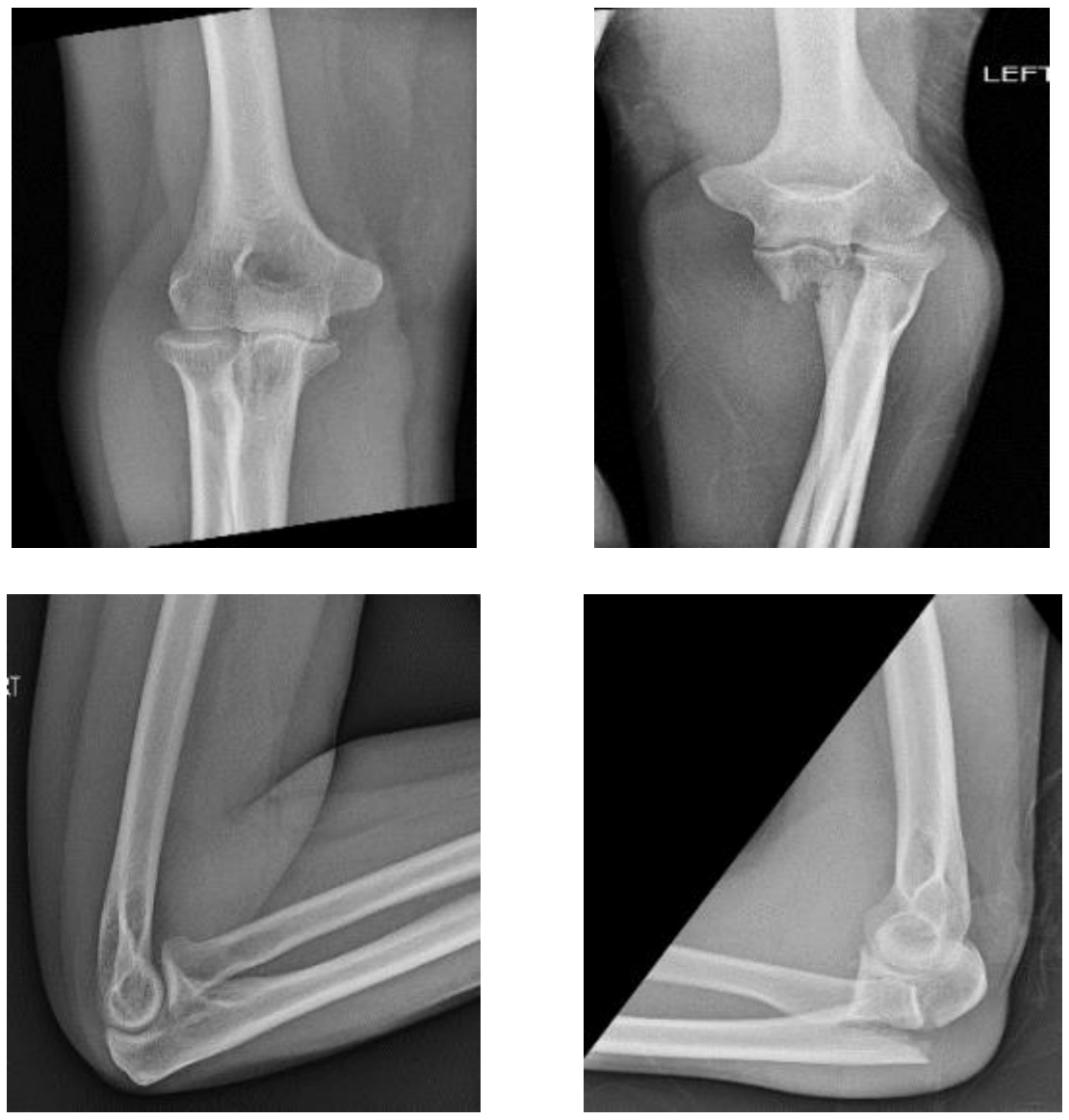

Figure 1(a): first row: AP view samples, hard (left)– easy (right) to diagnose; second row: Lateral view samples, hard (left) – easy (right) to diagnose.

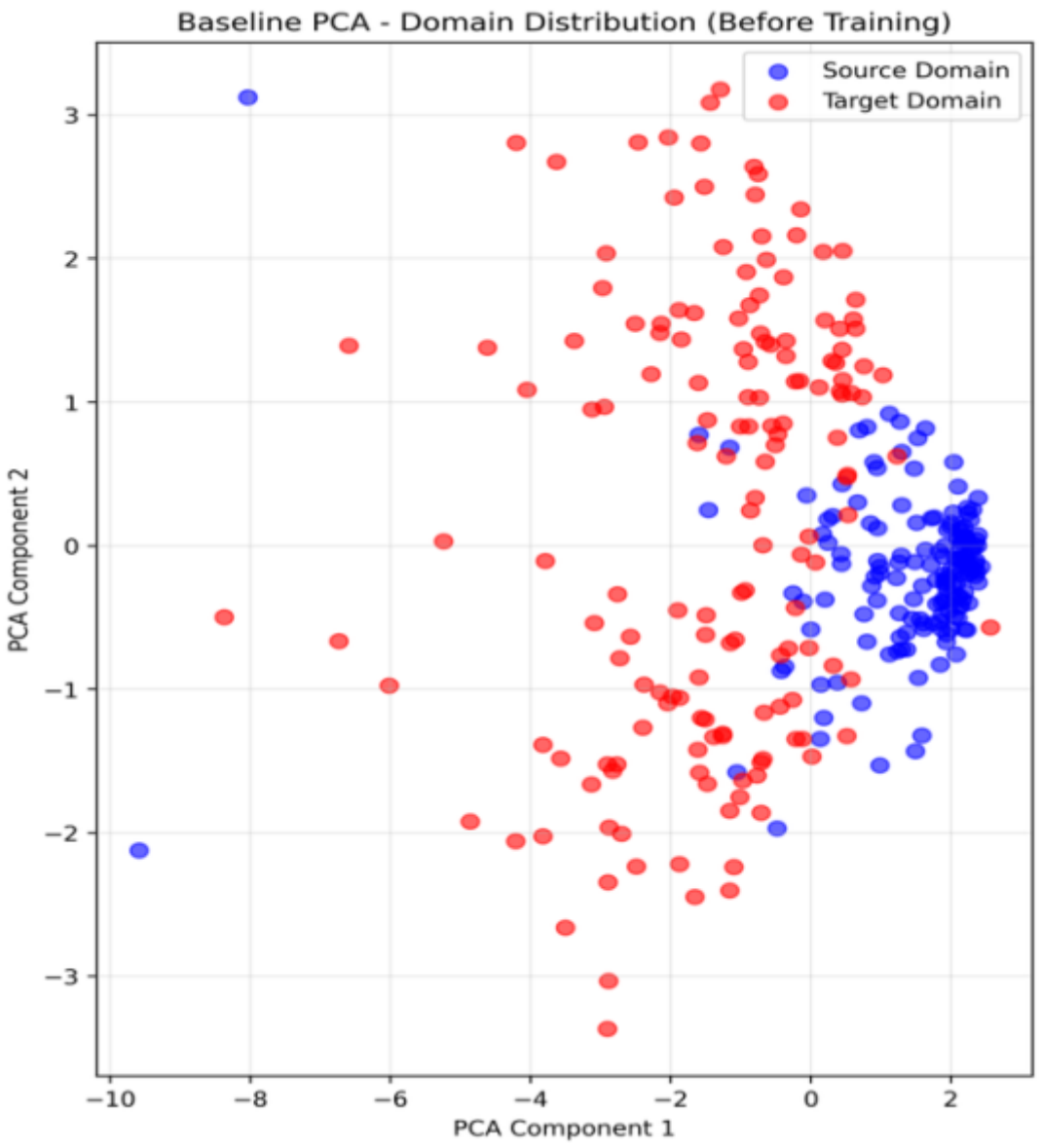

Figure 2(b): Two-dimensional PCA projection (PC1 vs. PC2) of deep features extracted by an ImageNet-pretrained encoder from identically preprocessed elbow radiographs: source = MURA (blue) and target = UofL (red). The clear separation between the two-point clouds indicates a substantial covariate (scanner/protocol) shift prior to training.

MURA [1] Dataset: MURA (**mu**sculoskeletal **ra**diographs) is a large dataset of bone X-rays, containing 14,863 musculoskeletal studies of the upper extremity [1]. Algorithms are tasked with determining whether an X-ray study is normal or abnormal. In this study we are focusing on Elbow category which contains 1912 studies, each contain several views. Figure 1(b) shows a baseline PCA of cross-domain features indicating a substantial covariate shift before training.

Table I: UofL and MURA Datasets per case

| Dataset | Training Set | | Test Set | | Total |
|---|---|---|---|---|---|
| | Normal | Fractured | Normal | Fractured | |
| UofL | 147 | 152 | 40 | 40 | 379 |
| MURA | 1094 | 660 | 92 | 66 | 1972 |

## IV. Method

We employ a domain-adversarial neural network in which a pretrained convolutional encoder (ResNet-50) is adapted to single-channel radiographs by replacing the first convolution to accept grayscale inputs and retaining global average pooling to produce a fixed-length feature vector (2048 for ResNet-50) [7]. A task classifier—an MLP with BatchNorm and Dropout—maps these features to fracture vs. normal logits. In parallel, a domain discriminator—another MLP—operates on the same features after a gradient-reversal layer (GRL) [2] to encourage domain-invariant representations. The GRL copies features forward but reverses the sign of the domain-loss gradient w.r.t the encoder, so the encoder learns representations that confuse the domain classifier while preserving fracture discriminations via the task-loss. To stabilize statistics, each training step encodes a concatenated mini batch of source and target images in a single forward pass and then splits the resulting features for task and domain heads, avoiding BatchNorm drift between domains.

Learning jointly optimizes discrimination and alignment. Let $\mathcal{L}_{cls}$ denote cross-entropy on labeled source samples and $\mathcal{L}_{dom}$ the cross-entropy of the domain discriminator applied to mixed (source +target) features passed through the GRL. We minimize $\theta_f, \theta_c, \theta_d$, where they are encoder, classifier, and discriminator parameters, respectively, and $\lambda(p) = \frac{2}{1+ e^{-10p}} - 1$ increases with training progress where $p \in [0,1]$. The domain loss is computed per sample and re-weighted to account for unequal source/target dataset sizes within a mini batch, which prevents the larger domain from dominating the adversarial signal. The GRL scales gradients flowing into the encoder, while λ(p) also scales the domain term in the total loss to avoid overwhelming the task objective early in training.

Optimization uses Adam with momentum 0.9 and the standard DANN learning-rate schedule $\eta(p) = \frac{0.01}{(1+10p)^{0.75}}$. Source and target batches have equal size (8/8) and receive identical preprocessing (grayscale conversion, 224×224 resize, intensity normalization) and light geometric augmentation to avoid leaking domain-specific cues.

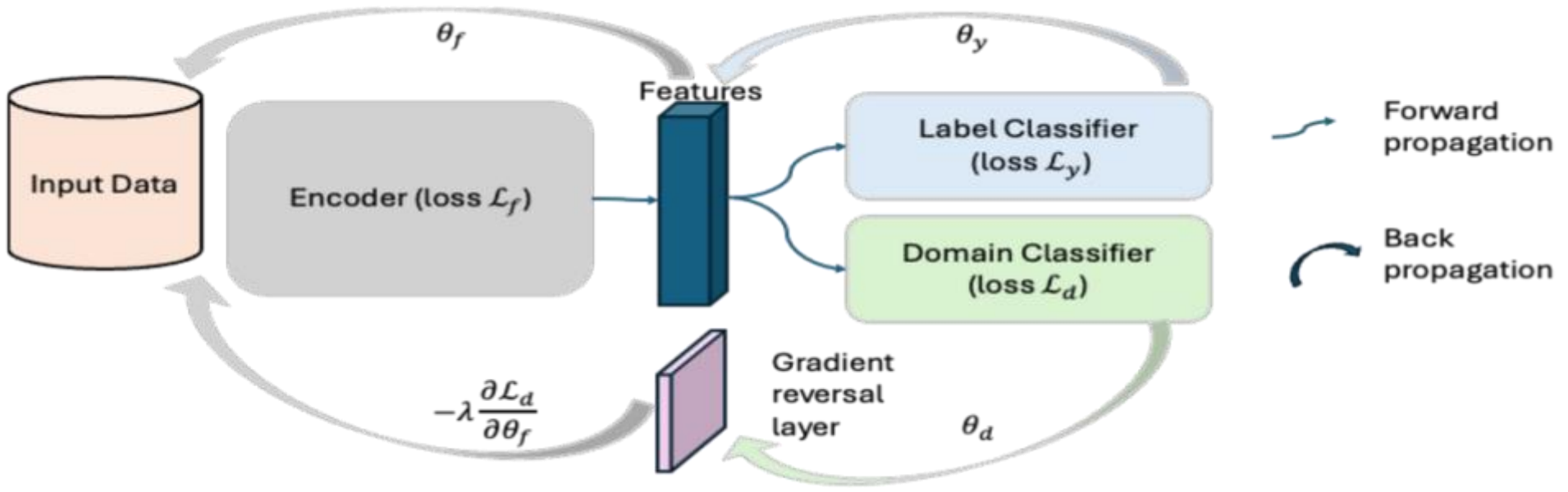

Figure 2 [2]: Adapted Domain Adversarial Neural Network (DANN) architecture [2]. The encoder (pretrained ResNet50) depicted in gray and label classifier in blue form a standard classification architecture. By adding Domain classifier (green) which is connected both to the feature vector and a gradient reversal layer multiplying the gradient by a negative hyperparameter during backpropagation suppressing domain loss ensuring the feature distributions over the two domains are as indistinguishable as possible.

Model selection is performed using target-domain validation AUC-ROC after each epoch, with early stopping and checkpointing of the best-performing weights on this criterion. For comparison, we implement three baselines under identical preprocessing and optimization: (1) Source-only training on MURA without the domain branch, (2) Target-only training on the small UofL set, and (3) Naive merge that pools MURA and UofL without adversarial alignment. The implementation is derived from a publicly available PyTorch DANN baseline and adapted to medical radiographs, ensuring reproducibility through fixed seeds, deterministic splits, and matched augmentation across domains.

## V. Results

We evaluated a pretrained ResNet-50 (grayscale-adapted) for adult elbow fracture classification under four training settings: Source-only (MURA), Target-only (UofL), Naive Merge (MURA + UofL), and DANN (MURA to UofL feature-level adaptation with gradient reversal). The final models were then tested on both held-out sets: UofL and MURA, shown in Table II.

**Tabel II**: Results For Models

| Method | AUC-ROC (UofL test set) | ACC (%) (UofL test set) | AUC-ROC (MURA test set) | ACC (%) (MURA test set) |
|---|---|---|---|---|
| Source-only | 0.647 | 59.331 | 0.905 | 83.79 |
| Target-only | 0.738 | 60.01 | 0.525 | 48.23 |
| Naive merge | 0.719 | 65.52 | 0.845 | 79.64 |
| EDANN (ours) | 0.831 | 80.65 | 0.893 | 82.88 |

## VI. Conclusion

For the full, finalized version, please refer to the proceedings of SPIE Medical Imaging 2026.